\documentclass[11pt,a4paper]{article}
\usepackage{authblk}
\usepackage[hyperref]{emnlp2018}
\usepackage{times}
\usepackage{latexsym}

\definecolor{icmldarkblue}{rgb}{0,0.08,0.45}
\hypersetup{
  colorlinks = true,
  citecolor = icmldarkblue,
  linkcolor = icmldarkblue
}

\usepackage[utf8]{inputenc} 
\usepackage[T1]{fontenc}    
\usepackage{nicefrac}       
\usepackage{microtype}      
\usepackage{url}
\usepackage{amsfonts}
\usepackage{amsmath}
\usepackage{amssymb}
\usepackage{amsthm}
\usepackage{mathrsfs}
\usepackage{tikz}
\usepackage{graphicx}
\usepackage{algpseudocode}
\usepackage{verbatim}
\usepackage{xcolor}
\usepackage{enumitem}
\usepackage{booktabs}
\usepackage{multirow}
\usepackage{todonotes}
\usepackage{footnote}
\usepackage{listings}
\usepackage[font=small]{caption}
\usepackage[linesnumbered,ruled]{algorithm2e}
\makesavenoteenv{tabular}

\definecolor{codeBackground}{rgb}{1.00,1.00,1.00}

\lstnewenvironment{python}[2]
{
  \lstset{
    title={#1},
    backgroundcolor=\color{codeBackground},
    numbers=left,
    stepnumber=1,
    basicstyle=\ttfamily\footnotesize,
    keywordstyle=\color{blue}\bfseries,
    showstringspaces=false,
    xleftmargin=10pt,
    language=python,
    frame=t,
  }
}
{}

\aclfinalcopy 

\newcommand{\eg}{\textit{e.g.}}
\newcommand{\ie}{\textit{i.e.}}
\newcommand{\cf}{\textit{c.f.}}
\newcommand{\etc}{etc.}

\newcommand{\p}{\mathbf{p}}
\newcommand{\q}{\mathbf{q}}

\newcommand{\raml}{RAML}
\renewcommand{\sc}{SwitchOut}

\newcommand{\expo}[1]{\exp{\left\{ #1 \right\}}}
\newcommand{\ABS}[1]{\left| #1 \right|}

\usepackage{slashbox}

\title{SwitchOut: an Efficient Data Augmentation Algorithm\\for Neural Machine Translation}

\author[*,1]{Xinyi Wang}
\author[*,1,2]{Hieu Pham}
\author[1]{Zihang Dai}
\author[1]{Graham Neubig}
\affil[ ]{\texttt{\{xinyiw1,hyhieu,dzihang,gneubig\}@cs.cmu.edu}}
\affil[1]{Language Technology Institute, Carnegie Mellon University, Pittsburgh, PA 15213}
\affil[2]{Google Brain, Mountain View, CA 94043}
\date{}

\begin{document}
\maketitle
{\let\thefootnote\relax\footnote{{*: Equal contributions.}}}
\begin{abstract}

In this work, we examine methods for data augmentation for text-based tasks such as neural machine translation~(NMT). 
We formulate the design of a data augmentation policy with desirable properties as an optimization problem, and derive a generic analytic solution.
This solution not only subsumes some existing augmentation schemes, but also leads to an extremely simple data augmentation strategy for NMT: randomly replacing words in \textit{both the source sentence and the target sentence} with other random words from their corresponding vocabularies.
We name this method \sc.
Experiments on three translation datasets of different scales show that \sc~yields consistent improvements of about 0.5 BLEU, achieving better or comparable performances to strong alternatives such as word dropout~\citep{pervasive_dropout}. Code to implement this method is included in the appendix.


\end{abstract}

\section{\label{sec:intro}Introduction and Related Work}

Data augmentation algorithms generate extra data points from the empirically observed training set to train subsequent machine learning algorithms. While these extra data points may be of lower quality than those in the training set, their quantity and diversity have proven to benefit various learning algorithms~\citep{cut_out,deep_speech_2}. In image processing, simple augmentation techniques such as flipping, cropping, or increasing and decreasing the contrast of the image are both widely utilized and highly effective~\citep{dense_net,wide_res_net}.

However, it is nontrivial to find simple equivalences for NLP tasks like machine translation, because even slight modifications of sentences can result in significant changes in their semantics, or require corresponding changes in the translations in order to keep the data consistent. In fact, indiscriminate modifications of data in NMT can introduce noise that makes NMT systems brittle~\citep{noise_break_nmt}.

Due to such difficulties, the literature in data augmentation for NMT is relatively scarce. To our knowledge, data augmentation techniques for NMT fall into two categories.
The first category is based on back-translation~\citep{back_translate_nmt,investigate_back_translation}, which utilizes monolingual data to augment a parallel training corpus.
While effective, back-translation is often vulnerable to errors in initial models, a common problem of self-training algorithms~\citep{chapelle2009semi}.
The second category is based on word replacements. For instance,~\citet{da_low_resource_nmt} propose to replace words in the target sentences with rare words in the target vocabulary according to a language model, and then modify the aligned source words accordingly. While this method generates augmented data with relatively high quality, it requires several complicated preprocessing steps, and is only shown to be effective for low-resource datasets. Other generic word replacement methods include word dropout~\citep{pervasive_dropout,variational_dropout_rnn}, which uniformly set some word embeddings to $0$ at random, and Reward Augmented Maximum Likelihood~(\raml;~\citet{raml}), whose implementation essentially replaces some words in the target sentences with other words from the target vocabulary.

In this paper, we derive an extremely simple and efficient data augmentation technique for NMT. \textbf{First,} we formulate the design of a data augmentation algorithm as an optimization problem, where we seek the data augmentation policy that maximizes an objective that encourages two desired properties: smoothness and diversity. This optimization problem has a tractable analytic solution, which describes a generic framework of which both word dropout and RAML are instances. \textbf{Second,} we interpret the aforementioned solution and propose a novel method: independently replacing words in \textit{both the source sentence and the target sentence} by other words uniformly sampled from the source and the target vocabularies, respectively. Experiments show that this method, which we name \sc, consistently improves over strong baselines on datasets of different scales, including the large-scale WMT 15 English-German dataset, and two medium-scale datasets: IWSLT 2016 German-English and IWSLT 2015 English-Vietnamese.

\section{\label{sec:method}Method}
\subsection{\label{sec:notations}Notations}
We use uppercase letters, such as $X$, $Y$, \etc, to denote random variables and lowercase letters such as $x$, $y$, \etc, to denote the corresponding actual values. Additionally, since we will discuss a data augmentation algorithm, we will use a hat to denote augmented variables and their values, \eg~$\widehat{X}$, $\widehat{Y}$, $\widehat{x}$, $\widehat{y}$, \etc~We will also use boldfaced characters, such as $\p$, $\q$, \etc, to denote probability distributions.

\subsection{\label{sec:data_aug}Data Augmentation}
We facilitate our discussion with a probabilistic framework that motivates data augmentation algorithms. With $X$, $Y$ being the sequences of words in the source and target languages (\eg~in machine translation), the canonical MLE framework maximizes the objective
\begin{equation*}
  \label{eqn:mle}
  J_\text{MLE}(\theta)
  	= \mathbb{E}_{x, y \sim \widehat{\p}(X, Y)} \left[ \log{\p_\theta(y | x)} \right].
\end{equation*}
Here $\widehat{\p}(X, Y)$ is the empirical distribution over all training data pairs $(x, y)$ and $\p_{\theta}(y | x)$ is a parameterized distribution that we aim to learn, \eg~a neural network. A potential weakness of MLE is the mismatch between $\widehat{\p}(X, Y)$ and the true data distribution $\p(X, Y)$. Specifically, $\widehat{\p}(X, Y)$ is usually a bootstrap distribution defined only on the observed training pairs, while $\p(X, Y)$ has a much larger support, \ie~the entire space of valid pairs. This issue can be dramatic when the empirical observations are insufficient to cover the data space.

In practice, data augmentation is often used to remedy this support discrepancy by supplying additional training pairs. 
Formally, let $\q(\widehat{X}, \widehat{Y})$ be the augmented distribution defined on a larger support than the empirical distribution $\widehat{\p}(X, Y)$. 
Then, MLE training with data augmentation maximizes
\begin{equation*}
  \label{eqn:data_aug}
  J_\text{AUG}(\theta)
    = \mathbb{E}_{\widehat{x}, \widehat{y} \sim \q(\widehat{X},\widehat{Y})}
      \left[ \log{\p_{\theta}(\widehat{y} | \widehat{x})} \right].
\end{equation*}
In this work, we focus on a specific family of $\q$, which depends on the empirical observations by
\begin{equation*}
\q(\widehat{X},\widehat{Y}) = \mathbb{E}_{x, y \sim \widehat{\p}(x, y)} \left[ \q(\widehat{X},\widehat{Y} | x, y) \right].
\end{equation*}
This particular choice follows the intuition that an augmented pair $(\widehat{x}, \widehat{y})$ that diverges too far from any observed data is more likely to be invalid and thus harmful for training. The reason will be more evident later.

\subsection{\label{sec:good_data_aug}Diverse and Smooth Augmentation}
Certainly, not all $\q$ are equally good, and the more similar $\q$ is to $\p$, the more desirable $\q$ will be. Unfortunately, we only have access to limited observations captured by $\widehat{\p}$. Hence, in order to use $\q$ to bridge the gap between $\widehat{\p}$ and $\p$, it is necessary to utilize some assumptions about $\p$. Here, we exploit two highly generic assumptions, namely:
\begin{itemize}
  \item \textbf{Diversity:} $\p(X, Y)$ has a wider support set, which includes samples that are more diverse than those in the empirical observation set.
  \item \textbf{Smoothness:} $\p(X, Y)$ is smooth, and similar $(x, y)$ pairs will have similar probabilities.
\end{itemize}
To formalize both assumptions, let $s(\widehat{x}, \widehat{y}; x, y)$ be a similarity function that measures how similar an augmented pair $(\widehat{x}, \widehat{y})$ is to an observed data pair $(x, y)$. Then, an ideal augmentation policy $\q(\widehat{X}, \widehat{Y} | x, y)$ should have two properties. First, based on the smoothness assumption, if an augmented pair $(\widehat{x}, \widehat{y})$ is more similar to an empirical pair $(x, y)$, it is more likely that $(\widehat{x}, \widehat{y})$ is sampled under the true data distribution $\p(X, Y)$, and thus $\q(\widehat{X}, \widehat{Y} | x, y)$ should assign a significant amount of probability mass to $(\widehat{x}, \widehat{y})$. Second, to quantify the diversity assumption, we propose that the entropy $\mathbb{H}[\q(\widehat{X}, \widehat{Y} | x, y)]$ should be large, so that the support of $\q(\widehat{X}, \widehat{Y})$ is larger than the support of $\widehat{\p}$ and thus is closer to the support $\p(X, Y)$. Combining these assumptions implies that 
$\q(\widehat{X}, \widehat{Y} | x, y)$ should maximize the objective
\begin{equation}
  \label{eqn:crit_q}
  \begin{aligned}
  J(\q; x, y)
	&= \mathbb{E}_{\widehat{x}, \widehat{y} \sim \q(\widehat{X}, \widehat{Y} | x, y)} \big[ s(\widehat{x}, \widehat{y}; x, y) \big] \\
    &+ \tau \mathbb{H}(\q(\widehat{X}, \widehat{Y} | x, y)),
  \end{aligned}
\end{equation}
where $\tau$ controls the strength of the diversity objective. The first term in~\eqref{eqn:crit_q} instantiates the smoothness assumption, which encourages $\q$ to draw samples that are similar to $(x, y)$. Meanwhile, the second term in~\eqref{eqn:crit_q} encourages more diverse samples from $\q$. Together, the objective $J(\q; x, y)$ extends the information in the ``pivotal'' empirical sample $(x, y)$ to a diverse set of similar cases.
This echoes our particular parameterization of $\q$ in Section \ref{sec:data_aug}.

The objective $J(\q; x, y)$ in~\eqref{eqn:crit_q} is the canonical maximum entropy problem that one often encounters in deriving a max-ent model~\citep{berger1996maximum}, which has the analytic solution: 
\begin{equation}
  \label{eqn:data_aug}
    \q^*(\widehat{x}, \widehat{y} | x, y)
    = \frac{\expo{s(\widehat{x}, \widehat{y}; x, y) / \tau}}{\sum_{\widehat{x}', \widehat{y}'} \expo{s(\widehat{x}', \widehat{y}'; x, y) / \tau}}
\end{equation}
Note that \eqref{eqn:data_aug} is a fairly generic solution which is agnostic to the choice of the similarity measure $s$.
Obviously, not all similarity measures are equally good. 
Next, we will show that some existing algorithms can be seen as specific instantiations under our framework. Moreover, this leads us to propose a novel and effective data augmentation algorithm.

\subsection{\label{sec:existingandnew}Existing and New Algorithms}
\paragraph{Word Dropout.} In the context of machine translation,~\citet{pervasive_dropout} propose to randomly choose some words in the source and/or target sentence, and set their embeddings to $0$ vectors. Intuitively, it regards every new data pair generated by this procedure as similar enough and then includes them in the augmented training set. Formally, word dropout can be seen as an instantiation of our framework with a particular similarity function $s(\hat{x}, \hat{y}; x, y)$ (see Appendix~\ref{sec:word_dropout_as_data_aug}).

\paragraph{\raml.} From the perspective of reinforcement learning, \citet{raml} propose to train the model distribution to match a target distribution proportional to an exponentiated reward.
Despite the difference in motivation, it can be shown (\cf~Appendix \ref{sec:raml_as_data_aug}) that \raml~can be viewed as an instantiation of our generic framework, where the similarity measure is $s(\widehat{x}, \widehat{y}; x, y) = r(\widehat{y}; y)$ if $\widehat{x} = x$ and $-\infty$ otherwise. Here, $r$ is a task-specific reward function which measures the similarity between $\widehat{y}$ and $y$.
Intuitively, this means that RAML only exploits the smoothness property on the target side while keeping the source side intact. 

\paragraph{SwitchOut.} After reviewing the two existing augmentation schemes, there are two immediate insights. Firstly, augmentation should not be restricted to only the source side or the target side. Secondly, being able to incorporate prior knowledge, such as the task-specific reward function $r$ in \raml, can lead to a better similarity measure. 

Motivated by these observations, we propose to perform augmentation in both source and target domains. For simplicity, we separately measure the similarity between the pair $(\widehat{x}, x)$ and the pair $(\widehat{y}, y)$ and then sum them together, \ie
\begin{equation}
  \label{eqn:sim_dec}
    s(\widehat{x}, \widehat{y}; x, y) / \tau
      \approx r_x(\widehat{x}, x) / \tau_{x} + r_y(\widehat{y}, y) / \tau_{y},
\end{equation}
where $r_x$ and $r_y$ are domain specific similarity functions and $\tau_{x}$, $\tau_{y}$ are hyper-parameters that absorb the temperature parameter $\tau$. This allows us to factor $\q^*(\widehat{x}, \widehat{y} | x, y)$ into:
\begin{equation}
  \label{eqn:data_aug_decompose}
  \begin{aligned}
    \q^*(\widehat{x}, \widehat{y} | x, y)
    &= \frac{\expo{r_x(\widehat{x}, x) / \tau_{x}}}{\sum_{\widehat{x}'} \expo{r_x(\widehat{x}', x) / \tau_{x}}} \\
    &\times \frac{\expo{r_y(\widehat{y}, y) / \tau_{y}}}{\sum_{\widehat{y}'} \expo{r_y(\widehat{y}', y) / \tau_{y}}}
  \end{aligned}
\end{equation}
In addition, notice that this factored formulation allows $\widehat{x}$ and $\widehat{y}$ to be sampled independently.

\paragraph{Sampling Procedure.} To complete our method, we still need to define $r_{x}$ and $r_{y}$, and then design a practical sampling scheme from each factor in \eqref{eqn:data_aug_decompose}. Though non-trivial, both problems have been (partially) encountered in \raml~\citep{raml,softmax_q_dist}. For simplicity, we follow previous work to use the \textit{negative} Hamming distance for both $r_{x}$ and $r_{y}$. For a more parallelized implementation, we sample an augmented sentence $\widehat{s}$ from a true sentence $s$ as follows:
\begin{enumerate}[leftmargin=*]
  \vspace{-2pt}
  \item Sample $\widehat{n} \in \{0, 1, ..., \ABS{s}\}$ by $\p(\widehat{n}) \propto e^{-\widehat{n} / \tau}$.
  \item For each $i \in \{1,2,...,\ABS{s}\}$, with probability $\widehat{n} / \ABS{s}$, we can replace $s_i$ by a uniform $\widehat{s_i}\neq s_i$.
  \vspace{-2pt}
\end{enumerate}
This procedure guarantees that any two sentences $\widehat{s}_1$ and $\widehat{s}_2$ with the same Hamming distance to $s$ have the same probability, but slightly changes the relative odds of sentences with different Hamming distances to $s$ from the true distribution by negative Hamming distance, and thus is an approximation of the actual distribution. However, this efficient sampling procedure is much easier to implement while achieving good performance. 

Algorithm~\ref{algo:switch_out_sampling} illustrates this sampling procedure, which can be applied independently and in parallel for each batch of source sentences and target sentences. Additionally, we open source our implementation in TensorFlow and in PyTorch (respectively in Appendix~\ref{sec:code_tf} and~\ref{sec:code_pytorch}).
\begin{algorithm}[h]
  \caption{\label{algo:switch_out_sampling}Sampling with \sc.}
  \small
  \DontPrintSemicolon
  \SetKwInOut{Input}{Input}
  \SetCommentSty{itshape}
  \SetKwComment{Comment}{$\triangleright$\ }{}
  \SetKwInOut{Output}{Output}
  \SetKwFunction{FSample}{HammingDistanceSample}
  \SetKwProg{Fn}{Function}{:}{}
  \SetKwProg{ParFor}{In parallel, do:}{}{}
  \Input{ $s$: a sentence represented by vocab integral ids,
          $\tau$: the temperature,
          $V$: the vocabulary}
  \Output{ $\widehat{s}$: a sentence with words replaced}
  \Fn{\FSample{$s$, $\tau$, $\ABS{V}$}}{
    Let $Z(\tau) \leftarrow \sum_{n=0}^{\ABS{s}} e^{-n / \tau}$ be the partition function.\;
    Let $p(n) \leftarrow e^{-n / \tau} / Z(\tau)$ for $n = 0, 1, ..., \ABS{s}$.\;
    Sample $\widehat{n} \sim p(n)$.\;
    \ParFor{}{
      Sample $a_i \sim \text{Bernoulli}(\widehat{n} / \ABS{s})$.\;
      \eIf{$a_i = 1$}{
        $\widehat{s}_i \leftarrow \text{Uniform}(V \backslash \{s_i\})$.\;
      }{
        $\widehat{s}_i \leftarrow s_i$.\;
      }
    }
    \KwRet $\widehat{s}$
  }
\end{algorithm}


\section{\label{sec:exp}Experiments}
\paragraph{Datasets.} We benchmark \sc~on three translation tasks of different scales: 1) IWSLT 2015 English-Vietnamese~(en-vi); 2) IWSLT 2016 German-English~(de-en); and 3) WMT 2015 English-German~(en-de). All translations are word-based. These tasks and pre-processing steps are standard, used in several previous works. Detailed statistics and pre-processing schemes are in Appendix~\ref{sec:datasets}.

\paragraph{Models and Experimental Procedures.} Our translation model, \ie~$\p_{\theta}(y | x)$, is a Transformer network~\citep{transformer}. For each dataset, we first train a standard Transformer model without~\sc~and tune the hyper-parameters on the dev set to achieve competitive results.
(w.r.t.~\citet{luong15iwslt,na_nmt,transformer}). Then, fixing all hyper-parameters, and fixing $\tau_{y} = 0$, we tune the $\tau_{x}$ rate, which controls how far we are willing to let $\widehat{x}$ deviate from $x$. Our hyper-parameters are listed in Appendix~\ref{sec:hparams}.

\paragraph{Baselines.} While the Transformer network without \sc~is already a strong baseline, we also compare \sc~against two other baselines that further use existing varieties of data augmentation: 1) word dropout on the source side with the dropping probability of $\lambda_\text{word} = 0.1$; and 2) RAML on the target side, as in Section~\ref{sec:existingandnew}. Additionally, on the en-de task, we compare \sc~against back-translation~\citep{back_translate_nmt}.

\paragraph{\sc~vs. Word Dropout and RAML.} We report the BLEU scores of \sc, word dropout, and RAML on the test sets of the tasks in Table~\ref{tab:results}. To account for variance, we run each experiment multiple times and report the median BLEU. Specifically, each experiment without \sc~is run for $4$ times, while each experiment with \sc~is run for $9$ times due to its inherently higher variance. We also conduct pairwise statistical significance tests using paired bootstrap~\citep{significance_test}, and record the results in Table~\ref{tab:results}.
For 4 of the 6 settings, \sc~delivers significant improvements over the best baseline without \sc. For the remaining two settings, the differences are not statistically significant. The gains in BLEU with \sc~over the best baseline on WMT 15 en-de are all significant~($p < 0.0002$). Notably, \sc~on the source demonstrates as large gains as these obtained by RAML on the target side, and \sc~delivers further improvements when combined with RAML.
\begin{table}[t!]
  \centering
  \resizebox{\columnwidth}{!}{%
  \begin{tabular}{lccc}
    \toprule
    \multicolumn{1}{l|}{\textbf{Method}}
    & \textbf{en-de}
    & \textbf{de-en}
    & \textbf{en-vi} \\
    \midrule
    \multicolumn{1}{l|}{Transformer}
    & 21.73
    & 29.81
    & 27.97 \\
    \multicolumn{1}{l|}{+WordDropout}
    & 20.63
    & \textbf{29.97}
    & 28.56 \\
    \multicolumn{1}{l|}{+\sc       }
    & \textbf{22.78${}^\dagger$}
    & 29.94
    & \textbf{28.67${}^\dagger$} \\
    \midrule
    \multicolumn{1}{l|}{+RAML}
    & 22.83
    & 30.66
    & 28.88 \\
    \multicolumn{1}{l|}{+RAML +WordDropout}
    & 20.69
    & 30.79
    & 28.86 \\
    \multicolumn{1}{l|}{+RAML +\sc}
    & \textbf{23.13${}^\dagger$}
    & \textbf{30.98${}^\dagger$}
    & \textbf{29.09} \\
    \bottomrule
  \end{tabular}
  }
  \caption{\label{tab:results}Test BLEU scores of \sc~and other baselines (median of multiple runs). Results marked with ${}^\dagger$ are statistically significant compared to the best result without \sc. For example, for en-de results in the first column, +\sc~has significant gain over Transformer; +RAML +\sc~has significant gain over +RAML.}
\end{table}

\paragraph{\sc~vs. Back Translation.} Traditionally, data-augmentation is viewed as a method to enlarge the training datasets~\citep{alex_net,inception_deeper}. In the context of neural MT,~\citet{back_translate_nmt} propose to use artificial data generated from a weak back-translation model, effectively utilizing monolingual data to enlarge the bilingual training datasets. In connection, we compare \sc~against back translation. We only compare \sc~against back translation on the en-de task, where the amount of bilingual training data is already sufficiently large\footnote{We add the extra monolingual data from \href{http://data.statmt.org/rsennrich/wmt16_backtranslations/en-de/}{http://data.statmt.org/rsennrich/wmt16\_backtranslations/en-de/}}. The BLEU scores with back-translation are reported in Table~\ref{tab:back_trans}. These results provide two insights. First, the gain delivered by back translation is less significant than the gain delivered by \sc. Second, \sc~and back translation are not mutually exclusive, as one can additionally apply \sc~on the additional data obtained from back translation to further improve BLEU scores.

\newcommand{\bt}{BT}
\begin{table}[t!]
  \centering
  \resizebox{0.70\columnwidth}{!}{%
  \begin{tabular}{lc}
    \toprule
    \multicolumn{1}{l|}{\textbf{Method}}
    & \textbf{en-de} \\
    \midrule
    \multicolumn{1}{l|}{Transformer}     & 21.73 \\
    \multicolumn{1}{l|}{+\sc}            & 22.78 \\
    \midrule
    \multicolumn{1}{l|}{+\bt}            & 21.82 \\
    \multicolumn{1}{l|}{+\bt~+RAML}      & 21.53 \\
    \multicolumn{1}{l|}{+\bt~+\sc}       & 22.93 \\
    \multicolumn{1}{l|}{+\bt~+RAML +\sc} & \textbf{23.76} \\
    \bottomrule
  \end{tabular}
  }
  \caption{\label{tab:back_trans}Test BLEU scores of back translation (\bt) compared to and combined with \sc~(median of $4$ runs).}
\end{table}

\paragraph{Effects of $\tau_x$ and $\tau_y$.} We empirically study the effect of these temperature parameters. During the tuning process, we translate the dev set of the tasks and report the BLEU scores in Figure~\ref{fig:raml_compare}. We observe that when fixing $\tau_{y}$, the best performance is always achieved with a non-zero $\tau_{x}$.
\begin{center}
\includegraphics[width=0.4\columnwidth]{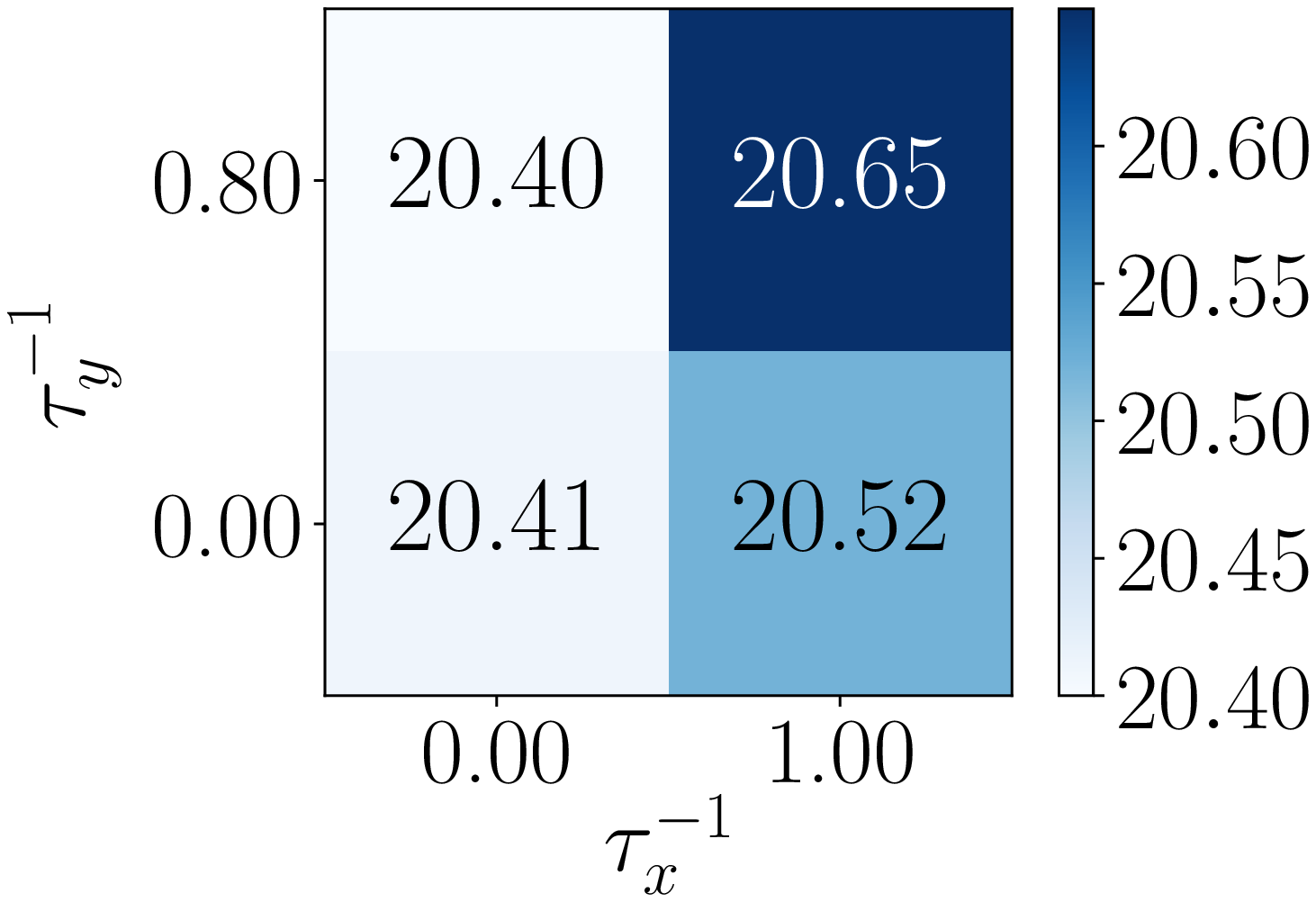}
  \includegraphics[width=0.55\columnwidth]{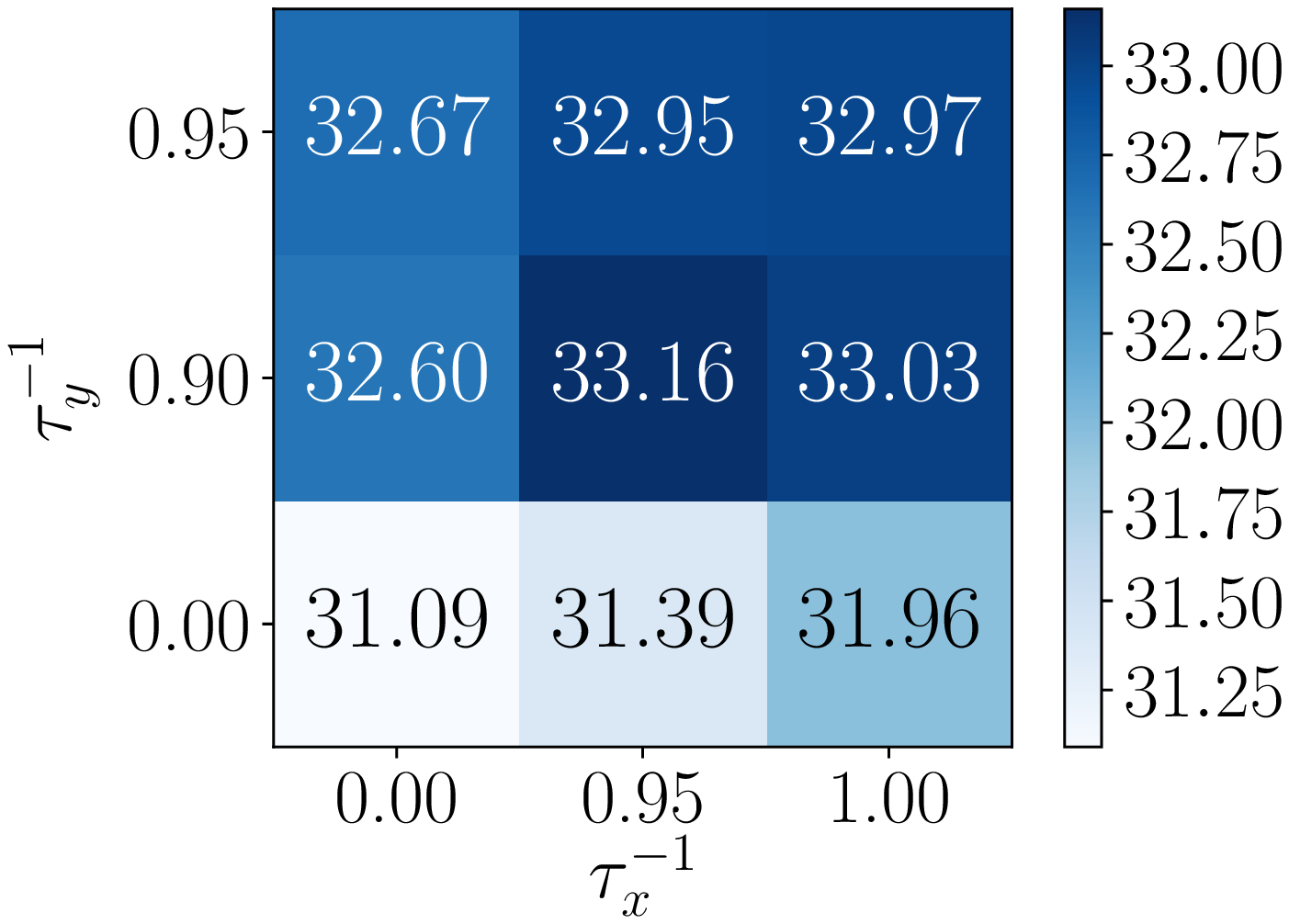}
  \includegraphics[width=0.55\columnwidth]{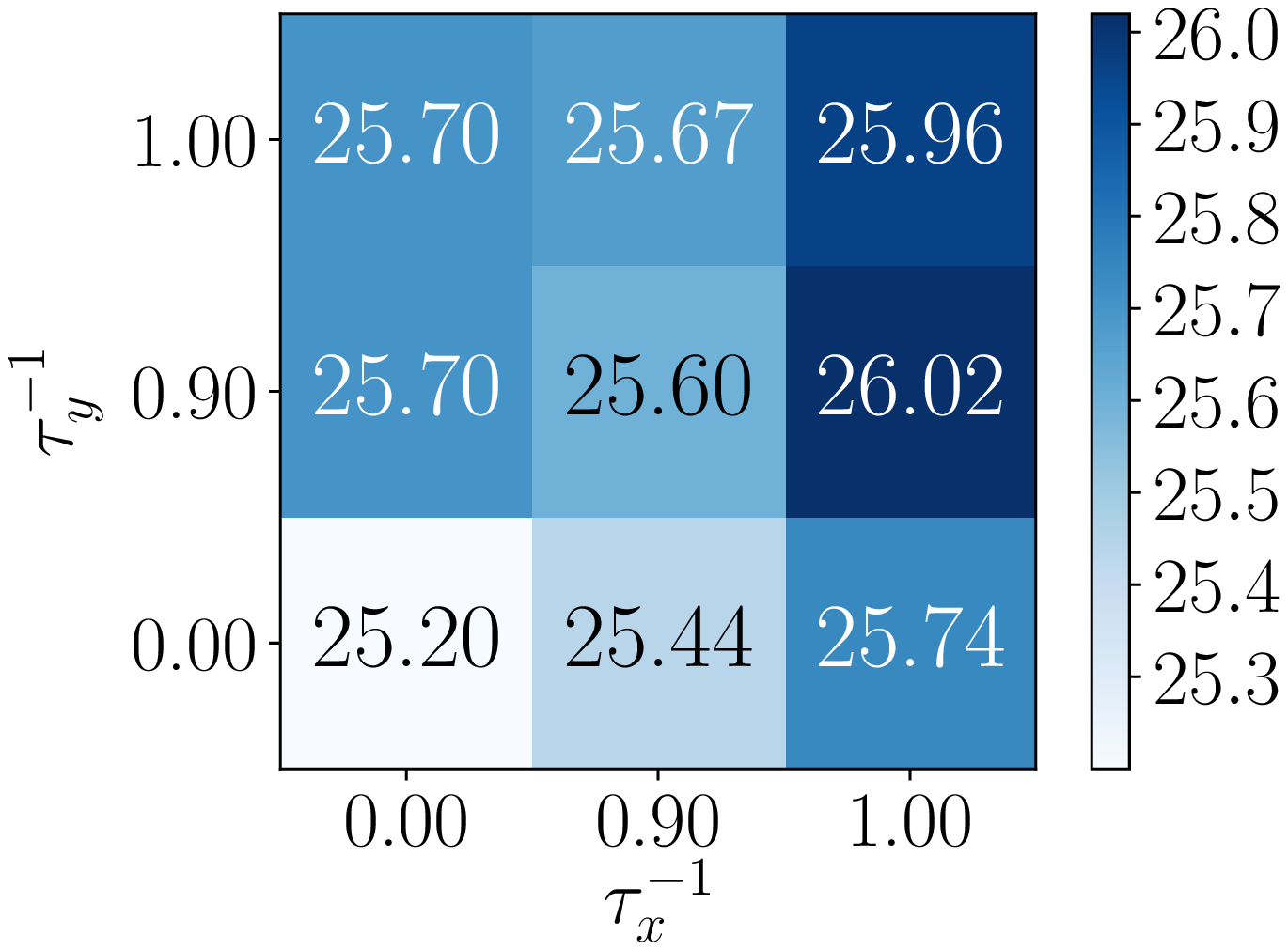}
  \captionof{figure}{\label{fig:raml_compare}Dev BLEU scores with different $\tau_{x}$ and $\tau_{y}$. \textit{Top left}: WMT 15 en-de. \textit{Top right}: IWSLT 16 de-en. \textit{Bottom}: IWSLT 15 en-vi.}
\end{center}
\paragraph{Where does~\sc~Help the Most?}
Intuitively, because~\sc~is expanding the support of the training distribution, we would expect that it would help the most on test sentences that are \emph{far from those in the training set} and would thus benefit most from this expanded support.
To test this hypothesis, for each test sentence we find its most similar training sample (\ie~nearest neighbor), then bucket the instances by the distance to their nearest neighbor and measure the gain in BLEU afforded by~\sc~for each bucket.
Specifically, we use (negative) word error rate~(WER) as the similarity measure, and plot the bucket-by-bucket performance gain for each group in Figure~\ref{fig:wer_plot}.
As we can see, \sc~improves increasingly more as the WER increases, indicating that \sc~is indeed helping on examples that are far from the sentences that the model sees during training. This is the desirable effect of data augmentation techniques.
\begin{center}
	\includegraphics[width=0.23\textwidth]{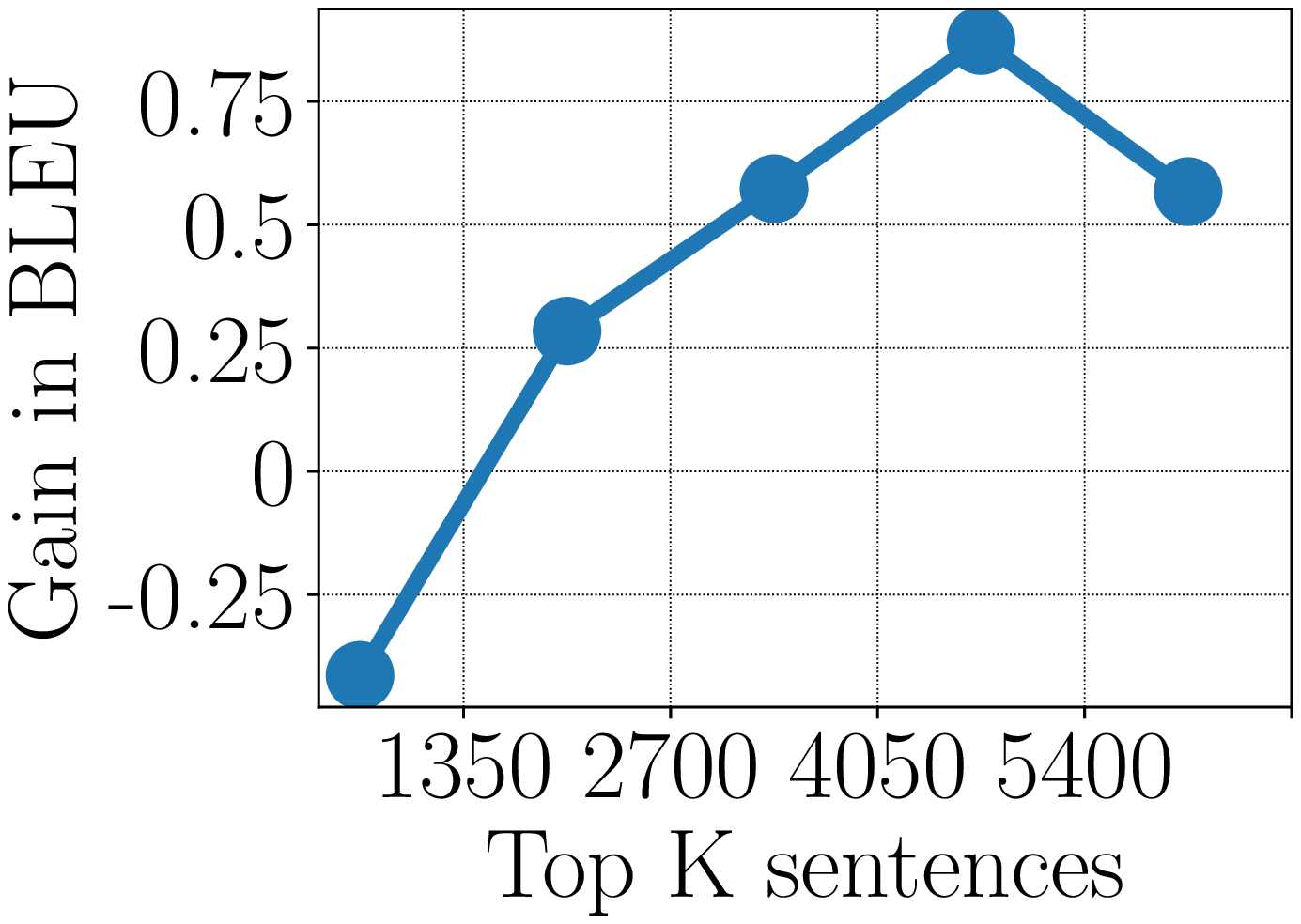}
  \includegraphics[width=0.23\textwidth]{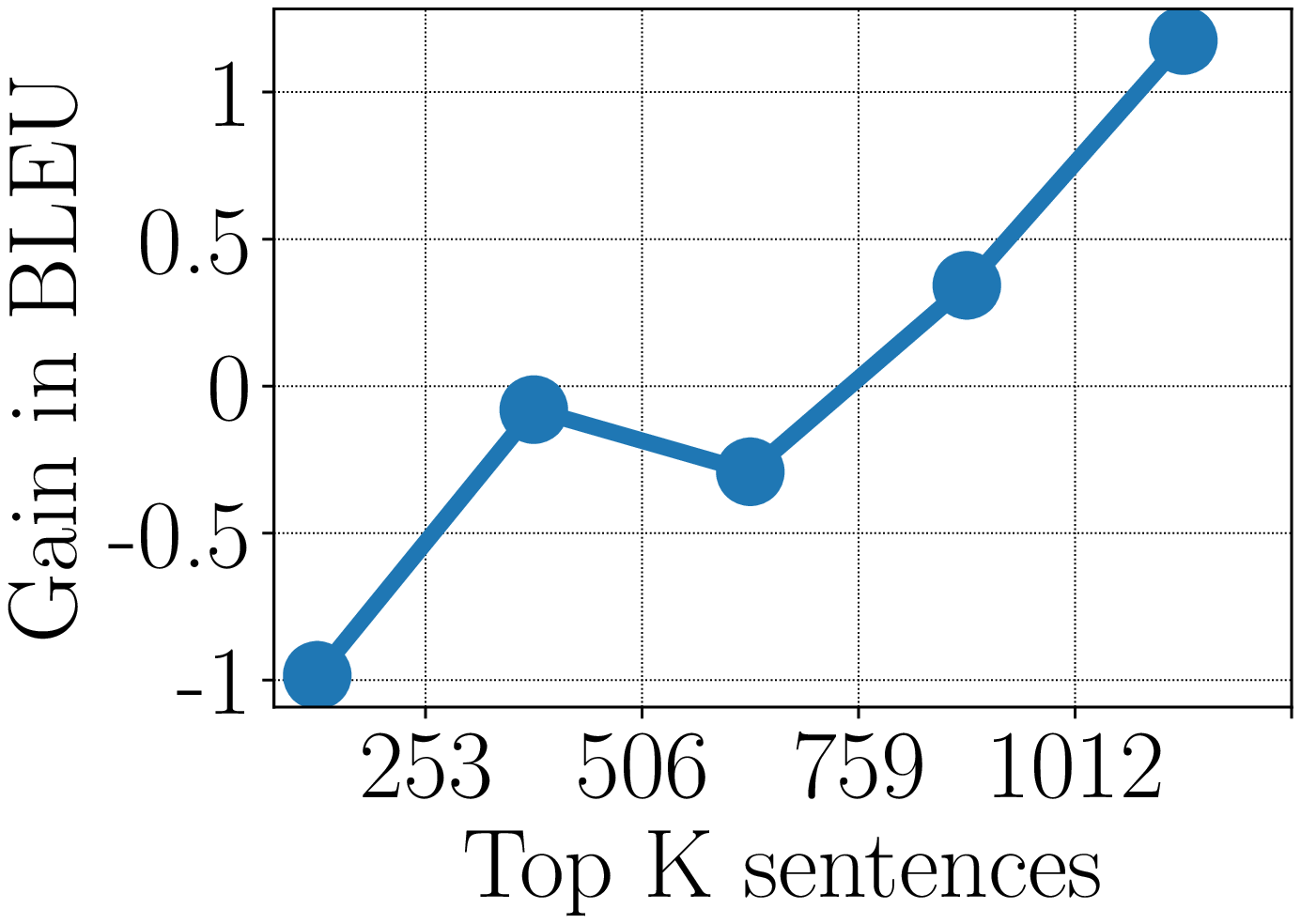}
    \captionof{figure}{\label{fig:wer_plot}Gains in BLEU of \raml+\sc~over \raml. $x$-axis is ordered by the WER between a test sentence and its nearest neighbor in the training set. \textit{Left}: IWSLT 16 de-en. \textit{Right}: IWSLT 15 en-vi.}
\end{center}


\section{\label{sec:conclusion}Conclusion}
In this paper, we propose a method to design data augmentation algorithms by solving an optimization problem. These solutions subsume a few existing augmentation schemes and inspire a novel augmentation method, \sc.~\sc~delivers improvements over translation tasks at different scales. Additionally,~\sc~is efficient and easy to implement, and thus has the potential for wide application.
\section*{Acknowledgements}

We thank Quoc Le, Minh-Thang Luong, Qizhe Xie, and the anonymous EMNLP reviewers, for their suggestions to improve the paper.

This material is based upon work supported in part by the Defense Advanced Research Projects Agency Information Innovation Office (I2O) Low Resource Languages for Emergent Incidents (LORELEI) program under Contract No. HR0011-15-C0114. The views and conclusions contained in this document are those of the authors and should not be interpreted as representing the official policies, either expressed or implied, of the U.S. Government. The U.S. Government is authorized to reproduce and distribute reprints for Government purposes notwithstanding any copyright notation here on.

\bibliography{main}
\bibliographystyle{acl_natbib_nourl}

\appendix
\onecolumn{
\section{\label{sec:appendix}Appendix}
\subsection{\label{sec:word_dropout_as_data_aug}Word Dropout as a Special Case}
Here, we derive word dropout as an instance of our framework. First, let us introduce a new token, $\langle \text{null} \rangle$, into both the source vocabulary and the target vocabulary. $\langle \text{null} \rangle$ has the embedding of a all-$0$ vector and is never trained. For a sequence $x$ of words in a vocabulary with $\langle \text{null} \rangle$, we define the \textit{neighborhood} $N(x)$ to be:
\begin{equation*}
  N(x) = \Big\{x': \ABS{x'} = \ABS{x} \text{ and }
               x'_i \in \{x_i, \langle \text{null} \rangle\}
               \Big\}
\end{equation*}
In other words, $N(x)$ consists of $x$ and all the sentences obtained by replacing a few words in $x$ by $\langle \text{null} \rangle$. Clearly, all augmented sentences $\widehat{x}$ that are sampled from $x$ using word dropout fall into $N(x)$.

In~\eqref{eqn:data_aug_decompose}, the augmentation policy $\q^*(\widehat{x}, \widehat{y} | x, y)$ was decomposed into two independent terms, one of which samples the augmented source sentence $\widehat{x}$ and the other samples the augmented target sentence $\widehat{y}$
\begin{equation*}
  \begin{aligned}
    \q^*(\widehat{x}, \widehat{y} | x, y)
    = \underbrace{\frac{\expo{r_x(\widehat{x}, x) / \tau_{x}}}{\sum_{\widehat{x}'} \expo{r_x(\widehat{x}', x) / \tau_{x}}}}_{\q(\widehat{x} | x)}
    \times \underbrace{\frac{\expo{r_y(\widehat{y}, y) / \tau_{y}}}{\sum_{\widehat{y}'} \expo{r_y(\widehat{y}', y) / \tau_{y}}}}_{\q(\widehat{y} | y)}
  \end{aligned}
\end{equation*}
Word dropout is an instance of this decomposition, where $r_y$ takes the same form with $r_x$, given by:
\begin{equation}
  \begin{aligned}
    r_x(\widehat{x}, x) &= \begin{cases}
      -\text{HammingDistance}(\widehat{x}, x) &\text{if $\widehat{x} \in N(x)$} \\
      -\infty &\text{otherwise}
    \end{cases},
  \end{aligned}
\end{equation}
where $\text{HammingDistance}(\widehat{x}, x) = \sum_{i=1}^{\ABS{x}} \mathbf{1}[\widehat{x}_i \neq x_i]$. To see this is indeed the case, let $h$ be the Hamming distance for $\widehat{x} \in N(x)$ and set $\lambda_{\text{word}} = \expo{-1 / \tau_{x}}$, then we have:
\begin{equation}
  \begin{aligned}
    \expo{r_x(\widehat{x}, x) / \tau_{x}}
      = \expo{-h / \tau_x}
      = \expo{-h \cdot \log{\frac{1}{\lambda_\text{word}}}}
      = \expo{h \cdot \log{\lambda_\text{word}}}
      = {\lambda_\text{word}}^{h},
  \end{aligned}
\end{equation}
which is precisely the probability of dropping out $h$ words in $x$, where each word is dropped with the distribution $\text{Bernoulli}(\lambda_\text{word})$.

The difference between word dropout and \sc~comes in the fact that $N(x)$ is much smaller than the support of $\widehat{x}$ that \sc~can sample from, which is $V^{\ABS{x}}$ where $V$ is the vocabulary. Word dropout concentrates all augmentation probability mass into $N(x)$ while \sc~spreads the mass into a larger support, leading to a larger entropy. Meanwhile, both word dropout and \sc~are exponentially less likely to diverge a way from $x$, ensuring the smoothness desiderata of a good data augmentation policy, as we discussed in Section~\ref{sec:good_data_aug}.

\subsection{\label{sec:raml_as_data_aug}\raml~as a Special Case}
Here, we present a detailed description of how \raml~is a special case of our proposed framework.
For each empirical observation $(x, y) \sim \widehat{\p}$, ~\raml~defines a reward aware target distribution $\p_\text{\raml}(Y | x, y)$ for the model distribution $\p_\theta(Y \mid x)$ to match. 
Concretely, the target distribution in~\raml~has the form
\begin{equation*}
\p_\text{\raml}(\widehat{y} | x, y) = \frac{\expo{r(\widehat{y}; y) / \tau}}{\sum_{\widehat{y}'} \expo{r(\widehat{y}'; y) / \tau}},
\end{equation*}
where $r$ is the task reward function.
With this definition,~\raml~amounts to minimizing the expected KL divergence between $\p_\text{\raml}$ and $\p_\theta$, \ie
\begin{equation*}
\begin{aligned}
&\min_{\theta} \mathbb{E}_{x, y \sim \widehat{\p}} \left[ \mathrm{KL}(\p_\text{\raml}(Y | x, y) \| \p_\theta(Y \mid x) \right] \\
\iff &
\max_{\theta} \mathbb{E}_{x, y \sim \widehat{\p}} \left[ \mathbb{E}_{\widehat{y} \sim \p_\text{\raml}(Y | x, y)} \left[ \log \p_\theta(\widehat{y} \mid x) \right] \right] \\
\iff &
\max_{\theta} \mathbb{E}_{\widehat{y} \sim \p_\text{\raml}(Y)} \left[ \log \p_\theta(\widehat{y} \mid x) \right],
\end{aligned}
\end{equation*}
where $\p_\text{\raml}(Y)$ is the marginalized target distribution, \ie~$\p_\text{\raml}(Y) = \mathbb{E}_{x, y \sim \widehat{\p}} \left[ \p_\text{\raml}(Y | x, y) \right]$.
Now, notice that $\p_\text{\raml}(Y)$ is a member of the augmentation distribution family in consideration (\cf~Section \ref{sec:data_aug}).
Specifically, it is equivalent to a data augmentation distribution where 
\begin{align}
 &&\q(\widehat{x}, \widehat{y} \mid x, y) 
&= \mathbf{1}[\widehat{x} = x] \cdot \p_\text{\raml}(\widehat{y} | x, y) \nonumber\\
\iff&&
\frac{\expo{s(\widehat{x}, \widehat{y}; x, y) / \tau}}{\sum_{\widehat{x}', \widehat{y}'} \expo{s(\widehat{x}', \widehat{y}'; x, y) / \tau}} 
&= \mathbf{1}[\widehat{x} = x] \cdot \frac{\expo{r(\widehat{y}; y) / \tau}}{\sum_{\widehat{y}'} \expo{r(\widehat{y}'; y) / \tau}} \nonumber\\
\iff&&
\label{eqn:raml_as_data_aug}
s(\widehat{x}, \widehat{y}; x, y) &=
\begin{cases}
r(\widehat{y}; y),& \widehat{x} = x \\
-\infty,& \widehat{x} \neq x 
\end{cases}.
\end{align}
The last equality reveals an immediate connection between~\raml~and our proposed framework. In summary,~\raml~can be seen as a special case of our data augmentation framework, where the similarity function is defined by \eqref{eqn:raml_as_data_aug}. Practically, this means~\raml~only consider pairs with source sentences from the empirical set for data augmentation.

\subsection{\label{sec:datasets}Datasets Descriptions}
\begin{table}[htb!]
  \centering
  \begin{tabular}{lccccc}
    \toprule
    &
    \multicolumn{2}{c}{\textbf{vocab (K)}} &
    \multicolumn{3}{c}{\textbf{\#sents}} \\
    \cmidrule(lr){2-3} \cmidrule(lr){4-6}
    & \textbf{src} & \textbf{tgt}
    & \textbf{train} & \textbf{dev} & \textbf{test} \\
    \midrule
    \multicolumn{1}{l|}{\textbf{en-vi}}
    & 17.2
    & 7.7
    & 133.3K
    & 1.6K
    & 1.3K \\
    \multicolumn{1}{l|}{\textbf{de-en}}
    & 32.0
    & 22.8
    & 153.3K
    & 7.0K
    & 6.8K \\
    \multicolumn{1}{l|}{\textbf{en-de}}
    & 50.0
    & 50.0
    & 4.5M
    & 2.7K
    & 2.2K \\
    \bottomrule
  \end{tabular}
  \caption{\label{tab:datasets}Statistics of the datasets.}
\end{table}
Table \ref{tab:datasets} summarizes the statistics of the datasets in our experiments. The WMT 15 en-de dataset is one order of magnitude larger than the IWSLT 16 de-en dataset and the IWSLT 15 en-vi dataset. For the en-vi task, we use the data pre-processed by~\citet{luong15iwslt}. For the en-de task, we use the data pre-processed by~\citet{dot_prod_attention}, with \textit{newstest2014} for validation and \textit{newstest2015} for testing. For the de-en task, we use the data pre-processed by~\citet{mixer_nmt}. 

\subsection{\label{sec:hparams}Hyper-parameters}
\begin{table}[htb!]
  \centering
  \begin{tabular}{l|ccccccccccc}
    \toprule
    \textbf{Task}
    & $n_{\text{layers}}$
    & $n_{\text{heads}}$
    & $d_{\text{k}}$, $d_{\text{v}}$
    & $d_{\text{model}}$
    & $d_{\text{inner}}$
    & init
    & clip
    & $\lambda_{\text{drop}}$
    & $\tau_{x}^{-1}$
    & $\tau_{y}^{-1}$
    & $\lambda_{\text{word}}$ \\
    \midrule
    \textbf{en-de} & $8$ & $6$ & $64$ & $512$ & $1024$ & $0.04$
                   & $25.0$ & $0.10$ & $1.00$ & $0.80$ & $0.1$ \\
    \textbf{de-en} & $8$ & $5$ & $64$ & $288$ &  $507$ & $0.035$
                   & $25.0$ & $0.25$ & $0.95$ & $0.90$ & $0.1$\\
    \textbf{en-vi} & $4$ & $4$ & $64$ & $256$ &  $384$ & $0.035$
                   & $20.0$ & $0.15$ & $1.00$ & $0.90$ & $0.1$\\
  \bottomrule
  \end{tabular}
  \vspace{0.1cm}
  \caption{\label{tab:hparams}Hyper-parameters for our experiments.}
\end{table}
The hyper-parameters used in our experiments are in Table~\ref{tab:hparams}. All models are initialized uniformly at random in the range as reported in Table~\ref{tab:hparams}. All models are trained with Adam~\citep{adam}. Gradients are clipped at the threshold as specified in  Table~\ref{tab:hparams}. For the WMT en-de task, we use the legacy learning rate schedule as specified by~\citet{transformer}. For the de-en task and the en-vi task, the learning rate is initially $0.001$, and is decreased by a factor of $0.97$ for every $1000$ steps, starting at step $8000$. All models are trained for 100,000 steps, during which one checkpoint is saved for each $2500$ steps and the final evaluation is performed on the checkpoint with lowest perplexity on the dev set.

Multiple GPUs are used for each experiment. For the de-en and the en-vi experiments, if we use $n$ GPUs, where $n \in \{1, 2, 4\}$, then we only perform $10^5 / n$ updates to the models' parameters. We find that this is sufficient to make the models converge.

\subsection{\label{sec:code_tf}Source Code for Sampling in TensorFlow}
\begin{python}{\texttt{Hamming distance sampling in TensorFlow}}

import tensorflow as tf
def hamming_distance_sample(sents, tau, bos_id, eos_id, pad_id, vocab_size):
  """Sample a batch of corrupted examples from sents.

  Args:
    sents: Tensor [batch_size, n_steps]. The input sentences.
    tau: temperature.
    vocab_size: to create valid samples.

  Returns:
    sents: Tensor [batch_size, n_steps]. The corrupted sentences.
  """

  # mask
  mask = [
      tf.equal(sents, bos_id),
      tf.equal(sents, eos_id),
      tf.equal(sents, pad_id),
  ]
  mask = tf.stack(mask, axis=0)
  mask = tf.reduce_any(mask, axis=0)

  # first, sample the number of words to corrupt for each sentence
  batch_size, n_steps = tf.unstack(tf.shape(sents))
  logits = -tf.range(tf.to_float(n_steps), dtype=tf.float32) * tau
  logits = tf.expand_dims(logits, axis=0)
  logits = tf.tile(logits, [batch_size, 1])
  logits = tf.where(mask,
                    x=tf.fill([batch_size, n_steps], -float("inf")), y=logits)

  # sample the number of words to corrupt at each sentence
  num_words = tf.multinomial(logits, num_samples=1)
  num_words = tf.reshape(num_words, [batch_size])
  num_words = tf.to_float(num_words)

  # <bos> and <eos> should never be replaced!
  lengths = tf.reduce_sum(1.0 - tf.to_float(mask), axis=1)

  # sample corrupted positions
  probs = num_words / lengths
  probs = tf.expand_dims(probs, axis=1)
  probs = tf.tile(probs, [1, n_steps])
  probs = tf.where(mask, x=tf.zeros_like(probs), y=probs)
  bernoulli = tf.distributions.Bernoulli(probs=probs, dtype=tf.int32)

  pos = bernoulli.sample()
  pos = tf.cast(pos, tf.bool)

  # sample the corrupted values
  val = tf.random_uniform(
      [batch_size, n_steps], minval=1, maxval=vocab_size, dtype=tf.int32)
  val = tf.where(pos, x=val, y=tf.zeros_like(val))
  sents = tf.mod(sents + val, vocab_size)

  return sents
\end{python}

\subsection{\label{sec:code_pytorch}Source Code for Sampling in PyTorch}
\begin{python}{\texttt{Hamming distance sampling in Pytorch}}
def hamming_distance_sample(sents, tau, bos_id, eos_id, pad_id, vocab_size):
  """
  Sample a batch of corrupted examples from sents.

  Args:
      sents: Tensor [batch_size, n_steps]. The input sentences.
      tau: Temperature.
      vocab_size: to create valid samples.
  Returns:
      sampled_sents: Tensor [batch_size, n_steps]. The corrupted sentences.
  """
  
  mask = torch.eq(sents, bos_id) | torch.eq(sents, eos_id) | torch.eq(sents, pad_id)
  lengths = mask.float().sum(dim=1)
  batch_size, n_steps = sents.size()
  # first, sample the number of words to corrupt for each sentence
  logits = torch.arange(n_steps)
  logits = logits.mul_(-1).unsqueeze(0).expand_as(
    sents).contiguous().masked_fill_(mask, -float("inf"))
  logits = Variable(logits)
  probs = torch.nn.functional.softmax(logits.mul_(tau), dim=1)
  num_words = torch.distributions.Categorical(probs).sample()

  # sample the corrupted positions.
  corrupt_pos = num_words.data.float().div_(lengths).unsqueeze(
    1).expand_as(sents).contiguous().masked_fill_(mask, 0)
  corrupt_pos = torch.bernoulli(corrupt_pos, out=corrupt_pos).byte()
  total_words = int(corrupt_pos.sum())
  # sample the corrupted values, which will be added to sents
  corrupt_val = torch.LongTensor(total_words)
  corrupt_val = corrupt_val.random_(1, vocab_size)
  corrupts = torch.zeros(batch_size, n_steps).long()
  corrupts = corrupts.masked_scatter_(corrupt_pos, corrupt_val)
  sampled_sents = sents.add(Variable(corrupts)).remainder_(vocab_size)

  return sampled_sents
\end{python}
\end{document}